\documentclass[10pt,twocolumn,twoside]{IEEEtran}

\usepackage{amssymb}
\usepackage{multirow}
\usepackage{booktabs}
\usepackage{amsthm}

\usepackage{amsthm}

\usepackage{empheq} 
\usepackage{mathtools}
\usepackage{xcolor}
\usepackage{mathrsfs}
\usepackage{amsfonts}
\usepackage{amsmath}
\usepackage{amssymb}
\usepackage{graphicx}
\usepackage{fancybox}
\usepackage{float}
\usepackage{cite}
\usepackage{enumitem}
\usepackage{textcomp}
\usepackage{gensymb}
\usepackage{multicol}
\usepackage{subfigure}
\usepackage{algorithm,algcompatible,amsmath}
\usepackage{bm}

\usepackage{xspace}

\newcommand{\virg}[1]{``#1''}
\DeclareMathOperator{\Tr}{Tr}

\newcommand{\Prob}[1]{\mathbb{P}\left( #1 \right)}

\newcommand{\E}[2][]{\ensuremath{\mathbb{E}_{#1}{\left[ #2 \right]}\xspace}}

\DeclarePairedDelimiterX{\infdivx}[2]{(}{)}{%
  #1\;\delimsize|\delimsize|\;#2%
}
\newcommand{\KL}[2]{\ensuremath{\mathrm{KL} \infdivx{#1}{#2}}\xspace}

\let\oldforall\forall
\let\forall\undefined
\DeclareMathOperator{\forall}{\oldforall}

\renewcommand\vec{\mathbf}
\newcommand{\sset}[1]{\left\{ #1 \right\}} 

\usepackage{tikz}
\usetikzlibrary{positioning,fit,calc,backgrounds}
\usetikzlibrary{decorations.markings}
\definecolor{bleudefrance}{rgb}{0.19, 0.55, 0.91}
\definecolor{plategray}{gray}{0.5}

\makeatother

\title{Modeling uncertainty for Gaussian Splatting}

\author{Luca Savant,~\IEEEmembership{Student Member,~IEEE},
        Diego Valsesia,~\IEEEmembership{Member,~IEEE},
        Enrico Magli,~\IEEEmembership{Fellow,~IEEE} 
\thanks{The authors are with Politecnico di Torino - Department of Electronics and Telecommunications,  Italy.  Email:{name.surname}@polito.it.

This publication is part of the project PNRR-NGEU which has received funding from the MUR - DM 352/2022. This work was partially supported by the European Union under the Italian National Recovery and Resilience Plan (NRRP) of NextGenerationEU, partnership on ``Telecommunications of the Future'' (PE00000001 - program ``RESTART'').
}  }

\begin{document}
\maketitle

\begin{abstract}
We present Stochastic Gaussian Splatting (SGS): the first framework for uncertainty estimation using Gaussian Splatting (GS). 
GS recently advanced the novel-view synthesis field by achieving impressive reconstruction quality at a fraction of the computational cost of Neural Radiance Fields (NeRF). However, contrary to the latter, it still lacks the ability to provide information about the confidence associated with their outputs. 
To address this limitation, in this paper, we introduce a Variational Inference-based approach that seamlessly integrates uncertainty prediction into the common rendering pipeline of GS. Additionally, we introduce the Area Under Sparsification Error (AUSE) as a new term in the loss function, enabling optimization of uncertainty estimation alongside image reconstruction. 
Experimental results on the LLFF dataset demonstrate that our method outperforms existing approaches in terms of both image rendering quality and uncertainty estimation accuracy. Overall, our framework equips practitioners with valuable insights into the reliability of synthesized views, facilitating safer decision-making in real-world applications.
\end{abstract}

\section{Introduction} \label{sec:intro}

Novel-view synthesis, the task of generating images of a scene from viewpoints not observed during data collection, is a fundamental problem in computer vision with numerous applications, including virtual reality, augmented reality, and robotics. Traditionally, this task has been addressed using methods such as Structure from Motion \cite{ullman1979interpretation}, also called multi-view stereo, which rely on geometric reconstruction techniques. However, recent advances in deep learning, particularly with the introduction of Neural Radiance Fields (NeRF), have revolutionized the field by enabling high-fidelity synthesis of novel views directly from the underlying scene representation.

NeRF \cite{mildenhall2021nerf} has recently enjoyed great success by representing a scene as a continuous volumetric function that maps 3D spatial coordinates and viewing directions to radiance values. By learning this function from a set of posed images, NeRF can generate photorealistic images from novel viewpoints. However, while NeRF achieves impressive results, its computational complexity and memory requirements limit its practicality for real-time applications. This is the focus of the emerging Gaussian Splatting (GS) technique \cite{kerbl20233d} which offers a more computationally efficient alternative to NeRF while maintaining high-quality novel-view synthesis. GS learns to approximate the radiance field by using a set of Gaussian kernels, enabling real-time rendering with competitive visual fidelity. 

At the same time, research in novel view synthesis has started addressing the problem of estimating the epistemic uncertainty in order to understand the reliability of the generated views. Indeed, any practical downstream task that involves taking actions in the real world (such as robotics and autonomous systems) must consider not only the newly synthesized views but also their corresponding uncertainties, in order to potentially discard too uncertain yet promising actions. This scenario was first addressed in the seminal work of Shen et al. \cite{shen2021stochastic}, where they proposed a deep architecture, based on NeRF, called S-NeRF to also estimate meaningful uncertainty maps for each generated view.

At the moment, GS lacks a mechanism for estimating uncertainty in the synthesized views.
In this paper, we seek to address this limitation by proposing a novel framework for uncertainty estimation in GS. We extend the traditional deterministic GS framework to incorporate stochasticity, allowing us to predict uncertainty alongside synthesized views. Our approach leverages Variational Inference (VI) to learn the parameters of the GS radiance field in a Bayesian framework, enabling us to accurately estimate uncertainty without sacrificing computational efficiency. 

We can summarize our novel contributions as follows:
\begin{itemize}
    \item we introduce a novel framework for uncertainty estimation in GS, called Stochastic Gaussian Splatting (SGS), enabling real-time synthesis of high-quality images with accurate uncertainty predictions;
    \item we propose a VI-based approach to learn the parameters of the GS radiance field, allowing us to incorporate uncertainty prediction seamlessly into the rendering pipeline. Moreover, we innovate this learning process by augmenting Empirical Bayes with a loss function dependent on the area under the sparsification curve;
    \item we demonstrate the effectiveness of our approach through experiments on the challenging LLFF dataset, showing significant improvements in both rendering quality and uncertainty estimation metrics compared to state-of-the-art methods.
\end{itemize}

\section{Background}
\subsection{NeRF and Gaussian Splatting}

In recent years, the Structure from Motion and novel view synthesis tasks have been revolutionized by novel techniques based on Neural Radiance Fields (NeRF), introduced in \cite{mildenhall2021nerf} and based on deep neural network architectures. Following the paradigm of implicit neural representation \cite{chen2019learning}, the NeRF network learns a mapping $\left(\mathbf{x} \in \mathbb{R}^3, \mathbf{d} \in \mathbb{S}^2\right) \to \left(c(\mathbf{x},\mathbf{d}), \sigma(\mathbf{x})\right)$ where $\mathbf{x}$ represents a 3D point, $\mathbf{d}$ a view direction, $c$ the color field, and $\sigma$ the density field. Using the Direct Volume Rendering Integral \cite{max1995optical, chandrasekhar2013radiative}, the physical radiance field $\mathbf{I}$ can be obtained from $c$ and $\sigma$, hence an image can be formed. This learning process can be supervised with the pixels from a number of available views, so that the network correctly resynthesizes the original images. After successful training, the network is also able to coherently synthesize novel views.

In an ever-growing literature around NeRF, GS \cite{kerbl20233d} distinguishes itself by achieving state-of-the-art visual quality while maintaining competitive training times and, importantly, allowing high-quality and real-time novel-view synthesis ($\geq 30 \text{fps}$ at 1080p resolution). 
In order to achieve such high inference speed, GS replaces the deep neural network with an elliptical basis function approximation. By carefully choosing the basis functions as elliptical Gaussian kernels, the EWA Volume Splatting technique \cite{zwicker2001ewa} can be used to achieve real-time view synthesis.

\subsection{Structure from Motion uncertainty estimation}

An emerging research direction for the Structure from Motion field is the quantification of the uncertainty of the synthesized novel-views or of the 3D radiance field itself. This task was first addressed by \cite{shen2021stochastic}, where it is cast as a Bayesian learning problem and solved with the Variational Inference (VI) framework \cite{jospin2022hands}, applied to NeRF. Subsequently, the same authors proposed an evolution of their method \cite{shen2022conditional}, which differs from the previous work by dropping the independence assumptions between the color and opacity fields, hence recovering a more complex stochastic dependency graph using the Conditional Normalizing Flows framework \cite{yang2019pointflow}. Following these works, \cite{wei2023fg} drops all independence assumptions by using a generative Flow-GAN model \cite{grover2018flowgan}. Another approach is addressed in \cite{goli2023bayes} where the Laplace Approximation framework \cite{jospin2022hands} is used. Finally, two similar works, \cite{shen2023estimating} and \cite{sunderhauf2023density}, focus on developing methods that actively estimate high uncertainty in spatial areas that are not covered by the input views, using VI and Ensemble Learning frameworks, respectively. It is worth noting that all these methods rely on the use use of NeRF, while, to the best of our knowledge, there is no work currently addressing the GS framework.

\subsection{Radiance Field and Direct Volume Rendering}
In this section, we recall the notation for volume rendering, which will be useful in the remainder of the paper. Following the optical derivation of \cite{max1995optical}, let's denote the physical light intensity field with $I$, which increases or decreases along a ray segment, parameterized by $t$, by interacting with particles in a particle-filled volume. These interactions are described using the density field $\sigma$ (also called the extinction coefficient) and the color field $c$ (also called the emission term), with the following Cauchy problem:

\begin{equation}\label{eq:differential}
  \begin{cases}
    &\frac{dI}{dt} = -c(t)\sigma(t) +\sigma(t)I(t) \quad \forall t \in \left[0, T\right]\\
    &I(T) = I_T
\end{cases}  
\end{equation}

whose solution value at $t=0$ is:
\begin{equation}\label{eq:integral}
    I(0) = I_T e^{-\int_{0}^{T} \sigma(t) dt} + \int_{0}^{T} e^{-\int_{0}^{T} \sigma(z) dz} c(t) \sigma(t) dt
\end{equation}

Reparametrizing the segment with $\vec{r}(t) = \vec{o} + t \vec{d}$, setting $I_T = 0$ (in the case of a dark background), and assuming an isotropic density field, the rendering integral is obtained:

\begin{equation}\label{eq:NeRF}
    I(\vec{o}, \vec{d}) = \int_{0}^{T} e^{-\int_{0}^{t} \sigma\left(\vec{r}(s)\right) ds} c\left(\vec{r}(t), \vec{d} \right) \sigma\left(\vec{r}(t)\right) dt
\end{equation}

From a given set of posed views, a set of corresponding posed pixels and colors can be extracted as $\mathcal{D} = \sset{\left(\left(\vec{o}_i, \vec{d}_i \right), y_i\right)}_i$, where $i$ is an index that runs over all the pixels of all the images in the training set, $\vec{o}_i$ is the camera center of the $i$-th pixel, $\vec{d}_i$ is the 3d spatial direction of the $i$-th pixel from its camera center and $y_i$ is the color of the $i$-th pixel, so that a reconstruction loss can be used to regress the color and density fields:

\begin{equation}\label{eq:lossrec}
    \mathcal{L}_{\textrm{rec}} = \sum_{i} \ell \left( y_i , I(\vec{o}_i, \vec{d}_i) \right)
\end{equation}

To synthesize a novel view, the integral in Eq. \eqref{eq:NeRF} must be evaluated by placing $\vec{o}$ at the camera origin and letting $\vec{d}$ vary for each pixel in the new image plane.

\section{Method}

In this section, we present the proposed method, called Stochastic Gaussian Splatting (SGS), to enable uncertainty quantification in the Gaussian Splatting framework.

\begin{figure*}[t]
    \centering
    \begin{tikzpicture}[
    basicnode/.style={minimum size = 9mm, align=right,},
    basicline/.style={thick, ->, draw=black},
    fromfixedline/.style={basicline, decoration={
            markings,
            mark={
                at position 0
                with {
                    \filldraw[black] circle [radius=0.5mm];
                }
            }
        },
        postaction=decorate},
    learnablevar/.style={basicnode, text=bleudefrance, text width = 12mm, inner sep=0.2cm},
    fixedvar/.style={basicnode,},
    stochvar/.style={basicnode, circle, draw=black, very thick,},
    myplate/.style={rectangle, draw=plategray, very thick, dashed, rounded corners=2mm, fill=gray!10, fill opacity=0.3},
    ]
    \node[learnablevar]     (Sigma)         at (0,0.8*1)        {$\bm{\Sigma}_k$};
    \node[learnablevar]     (muprior)       at (0,0.8*0)        {${\bm{\mu}}^*_k$};
    \node[learnablevar]     (Gammapriorsq)  at (0,0.8*-1)       {$\sqrt{{\bm{\Gamma}}^*_k}$};
    \node[learnablevar]     (alphaprior)    at (0,0.8*-2)       {${\alpha}^*_k$};
    \node[learnablevar]     (Pipriorsq)     at (0,0.8*-3)       {$\sqrt{{\Pi}^*_k}$};
    \node[learnablevar]     (cprior)        at (0,0.8*-4.5)     {${c}^*_{klm}$};
    \node[learnablevar]     (Xipriorsq)     at (0,0.8*-5.5)     {$\sqrt{{\Xi}^*_{klm}}$};
    
    \node[fixedvar]         (Gammaprior)    at (3,0.8*-1)       {${\bm{\Gamma}}^*_k$};
    \node[fixedvar]         (Piprior)       at (3,0.8*-3)       {${\Pi}^*_k$};
    \node[fixedvar]         (Xiprior)       at (3,0.8*-5.5)     {${\Xi}^*_{klm}$};
    
    \node[stochvar]         (mu)            at (6,0.8*-1)       {$\bm{\mu}_k$};
    \node[stochvar]         (alpha)         at (6,0.8*-3)       {$\alpha_k$};
    \node[stochvar]         (c)             at (6,0.8*-5.5)     {$c_{klm}$};
    
    \node[stochvar]         (G)             at (9,0.8*-1)       {$\mathcal{G}_k$};
    \node[stochvar]         (z)             at (12,0.8*-1)      {$z_k$};
    
    \node[stochvar]         (I)             at (15,0.8*-3)      {$I$};
    \node[fixedvar]         (Camera)        at (15,0.8*1)      {Camera};

    \node[text=black!60] (plateLMlabel) [below right=-0.2 cm and 0.5 cm of c] {$L,M$};
    \node[text=black!60] (plateKlabel) [below right=0 cm and 6 cm of c] {$K$};

    \begin{scope}[on background layer]
    \node [fit={(cprior)(Xipriorsq)(Xiprior)(c)(plateLMlabel)}] (tmp) {};
    \node [myplate, fit={(muprior)(Gammapriorsq)(alphaprior)(Pipriorsq)(cprior)(Xipriorsq)(Gammaprior)(Piprior)(Xiprior)(mu)(alpha)(c)(G)(Sigma)(z)(plateKlabel)(tmp)}] (plateK) {};
    \node [myplate, fit={(cprior)(Xipriorsq)(Xiprior)(c)(plateLMlabel)}] (plateLM)  {};
    \end{scope}

    \draw[fromfixedline] (muprior.east) -| (mu.north);
    \draw[fromfixedline] (alphaprior.east) -| (alpha.north);
    \draw[fromfixedline] (cprior.east) -| (c.north);
    
    \draw[fromfixedline] (Gammapriorsq) -- (Gammaprior);
    \draw[fromfixedline] (Pipriorsq) -- (Piprior);
    \draw[fromfixedline] (Xipriorsq) -- (Xiprior);
    
    \draw[fromfixedline] (Gammaprior) -- (mu);
    \draw[fromfixedline] (Piprior) -- (alpha);
    \draw[fromfixedline] (Xiprior) -- (c);
    
    \draw[basicline] (mu) -- (G);
    \draw[fromfixedline] (Sigma) -| (G);
    \draw[basicline] (G) -- (z);
    \draw[basicline] (z.east) -| (I.north);
    
    \draw[basicline] (alpha.east) -| (z.south);
    \draw[basicline] (c.east) -| (I.south);
    
    \draw[fromfixedline] (Camera.west) -| (z.north);
    
    \end{tikzpicture}
    \caption{Bayesian Network Graphical Model of SGS. Learnable variables are depicted in \textcolor{bleudefrance}{blue}, while stochastic variables are circled. The \virg{Camera} node represents both the spatial coordinate of the pixel $(\vec{o}, \vec{d})$ and the corresponding camera intrinsic and extrinsic parameters. \textcolor{plategray}{Gray} dashed rectangles are used for the plate notation, i.e. variables repetitions.}
    \label{fig:enter-label}
\end{figure*}
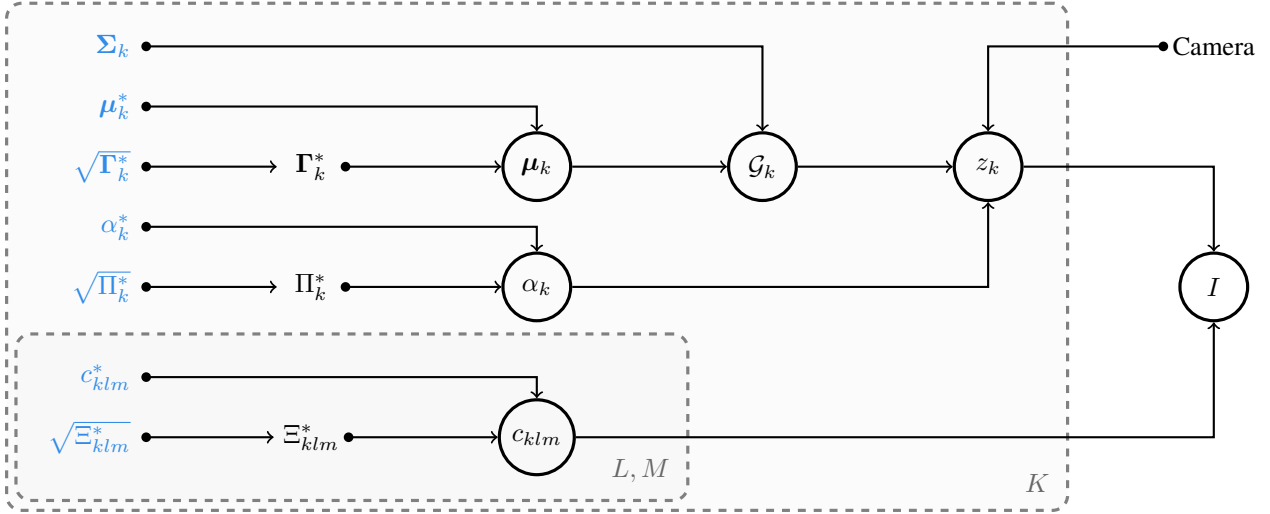

\subsection{Stochastic Gaussian Splatting}
The main difference between NeRF and GS lies in the fact that in the original NeRF formulation \cite{mildenhall2021nerf}, the color and density fields are regressed with a multilayer perceptron neural network. Instead, in GS \cite{kerbl20233d}, the two fields are learned with an elliptical basis function approximation. Let $\vec{x}$ be any point in 3D space, and let $\vec{d}$ identify a view direction, then the color and density fields are obtained as: 

\begin{equation}
    \begin{cases}
        \sigma(\vec{x}) \approx \sum_{k=1}^{K} \alpha_k \phi(\vec{x}; \bm{\mu}_k, \bm{\Sigma}_k) \\
        c(\vec{x},\vec{d}) \approx \sum_{k=1}^{K} c_k(\vec{d}) \phi(\vec{x}; \bm{\mu}_k, \bm{\Sigma}_k) 
    \end{cases};
\end{equation}

where:
\begin{itemize}
    \item $\phi$ is the elliptical 3D Gaussian kernel, with learnable parameters $\bm{\mu}_k \in \mathbb{R}^3$ and $\bm{\Sigma}_k \in \mathbb{R}^{3 \times 3}$ to control the shape and position of the $k$-th kernel:
        \begin{equation}
            \phi(\vec{x}; \bm{\mu}_k, \bm{\Sigma}_k) = \exp{\left(-\frac{1}{2} (\vec{x} - \bm{\mu}_k)^T \bm{\Sigma}_{k}^{-1} (\vec{x} - \bm{\mu}_k) \right)};
        \end{equation}
    \item $\alpha_k \in [0,1]$ is a learnable scalar to control the opacity of the $k$-th kernel;
    \item $c_k(\vec{d})$ is a learned linear combination of spherical harmonics $Y_{lm}$:
        \begin{equation}
            c_k(\vec{d}) = \sum_{l=0}^{L}\sum_{m=0}^{l} c_{klm} Y_{lm}(\vec{d}),
        \end{equation}
    where $l,m$ are degree and order of the used spherical harmonics.
\end{itemize}

In GS, the learnable parameters for each Gaussian kernel are $\bm{\mu}_k, \bm{\Sigma}_k, c_{klm}$, and $\alpha_k$.

The specific definition of $\phi$ enables the EWA Volume Splatting technique \cite{zwicker2001ewa}, so that the integral in Eq. \eqref{eq:NeRF} can be reduced to the simpler alpha blending technique:
\begin{equation}\label{eq:GS}
    I(\vec{o}, \vec{d}) = \sum_{k=1}^{K} \alpha_k e^{z_k(\vec{o}, \vec{d})} c_k(\vec{d}) \prod_{j=1}^{k-1} (1-\alpha_j e^{z_j(\vec{o}, \vec{d})})
\end{equation}
where $z_k(\vec{o}, \vec{d})$ is the splatting coefficient \cite{zwicker2001ewa}.

Aligning with previous works in the field, the proposed SGS method predicts an uncertainty value for each pixel in the novel-synthesized view. It is natural to define this uncertainty as the standard deviation of the predicted pixel color, requiring the predicted colors to be random variables. Hence, we need to inject stochasticity into the otherwise deterministic process of Eq. \eqref{eq:GS}. We propose to use a Monte Carlo method to approximate the variance of pixel colors, so that the expression \eqref{eq:GS} should be evaluated multiple times, each time by sampling some random variables.

Following the Variational Inference framework, we directly expand upon GS by imposing a prior distribution on each of the following parameters:
\begin{itemize}
    \item $\bm{\mu}_k \sim \mathcal{N}\left({\bm{\mu}}^*_k, {\bm{\Gamma}}^*_k\right)$
    \item $\ln \frac{\alpha_k}{1-\alpha_k} \sim \mathcal{N}\left({\alpha}^*_k, {\Pi}^*_k\right)$
    \item $c_{klm} \sim \mathcal{N}\left({c}^*_{klm}, {\Xi}^*_{klm}\right)$
\end{itemize}

so that the original GS parameters are no longer learned but are sampled from the above distributions, whose parameters are the new learning variables.

Thanks to the reparameterization trick \cite{kingma2013auto}, gradients can flow from the loss in Eq. \eqref{eq:lossrec}, which is a function of the samples, to the new distribution parameters ${\bm{\mu}}^*_k$, ${\bm{\Gamma}}^*_k$, ${\alpha}^*_k$, ${\Pi}^*_k$, ${c}^*_{klm}$ and ${\Xi}^*_{klm}$, enabling learning with standard backpropagation techniques.

\subsection{Learning with Variational Inference}
The Variational framework introduced in \cite{jospin2022hands} and used in \cite{shen2021stochastic} and \cite{shen2022conditional} is required for optimization, since direct optimization of just the loss would otherwise incur in underestimation of the pixel variance as the model could reduce to a deterministic one.

The VI framework stems from an approximation of Bayes' Theorem. Let $\mathcal{G}$ represent a Gaussian Splatting Radiance Field, composed of $K$ Gaussian kernels $\mathcal{G}_k$, and $\mathcal{D}$ represent the pixels dataset in Eq. \eqref{eq:lossrec}. Then the Bayes' Theorem reads as:
\begin{equation}
    \Prob{\mathcal{G} | \mathcal{D}} = \frac{\Prob{\mathcal{D} | \mathcal{G}} \Prob{\mathcal{G}}}{\Prob{\mathcal{D}}}.
\end{equation}
Since this is generally intractable, the VI framework prescribes to approximate the true posterior $\Prob{\mathcal{G} | \mathcal{D}}$ by introducing a parametric distribution $q_{\vec{\theta}}$ over all GS radiance fields $\mathcal{G}$ and to learn these parameters $\vec{\theta}$ in order to minimize the Kullback-Leibler (KL) divergence between the approximate posterior and the true one:

\begin{equation}\label{eq:yolo0}
    \min_{\vec{\theta}} \KL{q_{\vec{\theta}}(\mathcal{G})}{\Prob{\mathcal{G}|\mathcal{D}}}
\end{equation}

Now, this problem is further manipulated to get a tractable expression. Let us start with the following manipulation, where only properties of the logarithm and linearity on the expectation are used:

\begin{equation}\label{eq:yolo1}
\begin{aligned}
    &\KL{q_{\vec{\theta}}(\mathcal{G})}{\Prob{\mathcal{G}|\mathcal{D}}} = \\
    &= \E[q_{\vec{\theta}}]{\log \frac{q_{\vec{\theta}}(\mathcal{G})}{\Prob{\mathcal{G}|\mathcal{D}}}}
    = \E[q_{\vec{\theta}}]{\log \frac{q_{\vec{\theta}}(\mathcal{G}) \Prob{\mathcal{D}}}{\Prob{\mathcal{D}|\mathcal{G}} \Prob{\mathcal{G}}}}\\
    &= -\E[q_{\vec{\theta}}]{\log{\Prob{\mathcal{D}|\mathcal{G}}}} +\E[q_{\vec{\theta}}]{\log{\frac{q_{\vec{\theta}}(\mathcal{G})}{\Prob{\mathcal{G}}}}} + \E[q_{\vec{\theta}}]{\log{\Prob{\mathcal{D}}}} \\
    &= -\E[q_{\vec{\theta}}]{\log{\Prob{\mathcal{D}|\mathcal{G}}}} +\KL{q_{\vec{\theta}}(\mathcal{G})}{\Prob{\mathcal{G}}} + \log{\Prob{\mathcal{D}}}
\end{aligned}
\end{equation}

Now, remembering that $\mathcal{D} = \sset{\left(\left(\vec{o}_i, \vec{d}_i \right), y_i\right)}_i$, and that the dataset samples are independently sampled, the first term in the last expression is equivalent to:

\begin{align}\label{eq:yolo2}
    &\E[q_{\vec{\theta}}]{\log{\Prob{\mathcal{D}|\mathcal{G}}}} =\\
    &= \E[q_{\vec{\theta}}]{\log{\prod_i \Prob{\mathcal{D}_i|\mathcal{G}}}} \\
    &= \sum_i \E[q_{\vec{\theta}}]{\log{\Prob{\mathcal{D}_i|\mathcal{G}}}} \\
    &= \sum_i \E[q_{\vec{\theta}}]{\log{\Prob{\left(\left(\vec{o}_i, \vec{d}_i \right), y_i\right)|\mathcal{G}}}} \\
    &= \sum_i \E[q_{\vec{\theta}}]{\log{\Prob{y_i|\left(\vec{o}_i, \vec{d}_i \right),\mathcal{G}} \Prob{\left(\vec{o}_i, \vec{d}_i \right) | \mathcal{G}}}} \\
    &= \sum_i \E[q_{\vec{\theta}}]{\log{\Prob{y_i|\left(\vec{o}_i, \vec{d}_i \right),\mathcal{G}} \Prob{\vec{o}_i, \vec{d}_i }}} \\
    &= \sum_i \E[q_{\vec{\theta}}]{\log{\Prob{y_i|\left(\vec{o}_i, \vec{d}_i \right),\mathcal{G}}}} + \E[q_{\vec{\theta}}]{\log{\Prob{\vec{o}_i, \vec{d}_i }}} \\
    &= \sum_i \E[q_{\vec{\theta}}]{\log{\Prob{y_i|\left(\vec{o}_i, \vec{d}_i \right),\mathcal{G}}}} + \log{\Prob{\vec{o}_i, \vec{d}_i }}
\end{align}
Plugging \eqref{eq:yolo2} into \eqref{eq:yolo1}, we get:
\begin{equation}
\begin{aligned}
    &\KL{q_{\vec{\theta}}(\mathcal{G})}{\Prob{\mathcal{G}|\mathcal{D}}} =\\
    &= - \sum_i \E[q_{\vec{\theta}}]{\log{\Prob{y_i|\left(\vec{o}_i, \vec{d}_i \right),\mathcal{G}}}} + \\
    & - \sum_i \log{\Prob{\vec{o}_i, \vec{d}_i }} + \\
    & +\KL{q_{\vec{\theta}}(\mathcal{G})}{\Prob{\mathcal{G}}} + \log{\Prob{\mathcal{D}}}
\end{aligned}
\end{equation}
As the optimization variable is the parameters $\theta$, all the terms that do not depend on $\theta$ in the latter equation can be discarded. Finally, we have the optimization problem:

\begin{equation}\label{eq:KL2}
    \min_{\vec{\theta}} -\sum_{i} \E[q_{\vec{\theta}}]{\log \Prob{y_i|\left(\vec{o}_i, \vec{d}_i \right), \mathcal{G}}} + \KL{q_{\vec{\theta}}(\mathcal{G})}{\Prob{\mathcal{G}}}
\end{equation}

The first term in the loss function of problem \eqref{eq:KL2} is the expected negative log-likelihood. This term forces $\vec{\theta}$ to maximize the expected log-likelihood by matching the observations in the dataset $\mathcal{D}$. So, in spirit, it replaces the standard loss \eqref{eq:lossrec}. It is estimated with the Monte Carlo method and, for simplicity, by defining the conditional probability distribution $\Prob{y_i|\left(\vec{o}_i, \vec{d}_i \right), \mathcal{G}}$ to be a normal distribution that is pixel-wise independent, i.e.:
\begin{align}
    \log \Prob{y_i|\left(\vec{o}_i, \vec{d}_i \right), \mathcal{G}} \propto \left( y_i - I\left(\vec{o}_i, \vec{d}_i \right)\right)^2
\end{align}
Note that this requirement is not too strict as the independence is required in the conditional probability distribution and not for the unconditional probability distribution.

The second term in Eq. \eqref{eq:KL2}, instead, limits $\vec{\theta}$ from moving too far away from the prior distribution of the GS radiance field $\mathcal{G}$. For it to be efficiently tractable, we suppose independence among the $K$ Gaussian kernels, for both the prior and the approximate posterior distributions, i.e.:

\begin{align}
    q_\vec{\theta}(\mathcal{G}) &= \prod_{k=1}^{K} q_{\theta_k}(\mathcal{G}_k) \label{eq:indip} \\
    \Prob{\mathcal{G}} &= \prod_{k=1}^{K} \Prob{\mathcal{G}_k} \label{eq:indip2}
\end{align}

This independence assumption between Gaussian kernels is more general with respect to previous work. For instance, Shen et al. \cite{shen2021stochastic} prescribed independence between the values of the opacity fields for every pair of points in the radiance field, even if they are very close to each other. Meanwhile, in the proposed method, the independence is prescribed at the Gaussian kernel level and not at the infinitesimal 3D point level.

Thanks to the assumptions in Eqs. \eqref{eq:indip} and \eqref{eq:indip2}, the second term in Eq. \eqref{eq:KL2} can be expanded to become:

\begin{equation}\label{eq:KL3}
    \KL{q_{\vec{\theta}}(\mathcal{G})}{\Prob{\mathcal{G}}} = \sum_{k=1}^{K} \KL{q_{\theta_k}(\mathcal{G}_k)}{\Prob{\mathcal{G}_k}}
\end{equation}

If we suppose that the prior distributions are themselves multivariate Normal, each KL term in the right-hand side of Eq. \eqref{eq:KL3} has the following general closed form expression:

\begin{align*}
    \KL{\mathcal{N}_{\mu_0, \Sigma_0}}{\mathcal{N}_{\mu_1, \Sigma_1}} = &-\frac{3}{2} +\frac{1}{2}\Tr\left(\Sigma_1^{-1} \Sigma_0\right) \\
        &+\frac{1}{2}(\mu_1-\mu_0)^{T} \Sigma_1^{-1}(\mu_1-\mu_0) \\
        &+\frac{1}{2}\ln\left(\frac{\det \Sigma_1}{\det \Sigma_0}\right)
\end{align*}

Hence, we define the total KL contribution to the training loss $\mathcal{L}_{\textrm{KL}}$ as the sum over all Gaussians and over all the learnable parameters of the KL divergence between the prior (hat variables) and the posterior (starred variables) of that parameter:

\begin{align}
    \mathcal{L}_{\textrm{KL}} = \sum_{k=1}^{K} \Big[
       &\KL{\mathcal{N}_{\hat{\bm{\mu}}_k, \hat{\bm{\Gamma}}_k}}{\mathcal{N}_{{\bm{\mu}}^*_k, {\bm{\Gamma}}^*_k}} \nonumber
     \\ +&\KL{\mathcal{N}_{\hat{\alpha}_k, \hat{\Pi}_k}}{\mathcal{N}_{{\alpha}^*_k, {\Pi}^*_k}} \nonumber
     \\ +&\KL{\mathcal{N}_{\hat{c}_{klm}, \hat{\Xi}_{klm}}}{\mathcal{N}_{{c}^*_{klm}, {\Xi}^*_{klm}}} \Big] \label{eq:lkl}
\end{align}

\subsection{Learning with AUSE}

To assess the accuracy of uncertainty estimation, a quantitative approach involves examining its correlation with the true error map using the Sparsification Curve \cite{gustafsson2020evaluating}. First, the predicted values are sorted based on decreasing predicted uncertainty and then progressively removed, starting from those with high predicted uncertainty. By keeping track of a quality metric applied to the remaining values, the Sparsification Curve is generated. The area under this curve is called Area Under Sparsification Curve (AUSC). The Area Under the Sparsification Error (AUSE) metric is defined as the difference between the AUSC of the method and the AUSC of the oracle, i.e., the curve obtained by sorting the predicted values according to the true error.

If the uncertainty prediction was random, the percolation process would also be random, resulting in a flat curve and a high AUSE. Otherwise, if the uncertainty prediction was positively correlated with the prediction error, improvements in the tracked quality metric would be observed. As the GS technique has significantly lower memory requirements compared to the NeRF used in previous works, it is capable of sampling the whole view multiple times in a single forward pass. This enables us to directly compute the AUSE metric applied to all the pixels in a view, taking the standard deviation of the samples of each pixel as the uncertainty map.

As we will show in the experiments section, using only the standard VI framework is suboptimal, as VI would not fully exploit the fact that the GS technique is orders of magnitude faster than neural-network-based NeRFs. Hence, in this work, we exploit the efficiency and low memory requirements of GS by augmenting the VI loss of Eq. \eqref{eq:yolo0} with the AUSE metric.

\subsection{End-to-end SGS Training}\label{sec:learnprior}

We now have all the necessary ingredients to define the overall loss used to train the proposed SGS method. In particular, the overall loss is the following combination:
\begin{equation}\label{eq:lossfinal}
    \mathcal{L} = \mathcal{L}_{\textrm{rec}} + \lambda_{\textrm{SSIM}} \mathcal{L}_{\textrm{ssim}}  + \lambda_{\textrm{KL}} \mathcal{L}_{\textrm{KL}} + \lambda_{\textrm{AUSE}} \mathcal{L}_{\textrm{AUSE}} 
\end{equation}
where $\mathcal{L}_{\textrm{rec}}$ is defined as in \eqref{eq:lossrec} using the $\ell_1$ norm, $\mathcal{L}_{\textrm{ssim}}$ is the SSIM perceptual loss, $\mathcal{L}_{\textrm{KL}}$ is the Kullback–Leibler divergence with the prior from Eq. \eqref{eq:lkl}, and $\mathcal{L}_{\textrm{AUSE}}$ is the loss induced by the AUSE RMSE metric. The $\ell_1$, SSIM and AUSE losses augment the conventional KL loss in order to more explicitly enforce the training tradeoff between distortion, perceptual quality and uncertainty estimation.

Finally, one more aspect in which SGS training differs from previous works is the approach to learning the distribution of the priors. In previous works, stochasticity was introduced in the weights of neural networks, which are typically randomly initialized \cite{narkhede2022review}. However, in GS, the parameters have a more direct physical meaning in the 3D space. For example, minimizing the KL-divergence in GS would tend to fix the center of a Gaussian kernel fixed in a randomly initialized position in 3D space. Instead, we tackle this convergence issue, by taking inspiration from Empirical Bayes. We thus first learn an informative prior with some iterations of classic GS, which serves as an initialization before switching to the SGS formulation.

\section{Experimental results}
\begin{table*}[t]
\centering
\caption{Results on the LLFF dataset. * denotes extra depth information.}
\label{tab:results}
\begin{tabular}{lccccc}
                                                    & \multicolumn{3}{c}{Rendering Metrics} & \multicolumn{2}{c}{Uncertainty Metrics} \\ \hline
                                                    & PSNR $\uparrow$      & SSIM $\uparrow$ & LPIPS $\downarrow$ & AUSE RMSE $\downarrow$  & AUSE MAE $\downarrow$  \\ \hline
D.E. (Lakshminarayanan, Pritzel, and Blundell 2017) & \underline{22.32}      & 0.788      & 0.236     & 0.0254              & 0.0122 \\
Drop. (Gal and Ghahramani 2016)                     & 21.90      & 0.758      & 0.248     & 0.0316              & 0.0162            \\
NeRF-W (Martin-Brualla et al. 2021)                 & 20.19      & 0.706      & 0.291     & 0.0268              & 0.0113            \\
S-NeRF* (Shen et al. 2021)                           & 20.27      & 0.738      & 0.229     & 0.0248              & 0.0101            \\
CF-NeRF* (Shen et al. 2022)                          & 21.96      & \underline{0.790} & \underline{0.201} & \underline{0.0177} & \textbf{0.0078} \\
SGS (Ours)                                          & \textbf{24.20} & \textbf{0.842} & \textbf{0.121} & \textbf{0.0147} & \underline{0.0092} \\ \hline \\
\end{tabular}
\end{table*}

\begin{figure*}[t]
    \includegraphics[width=0.98 \textwidth]{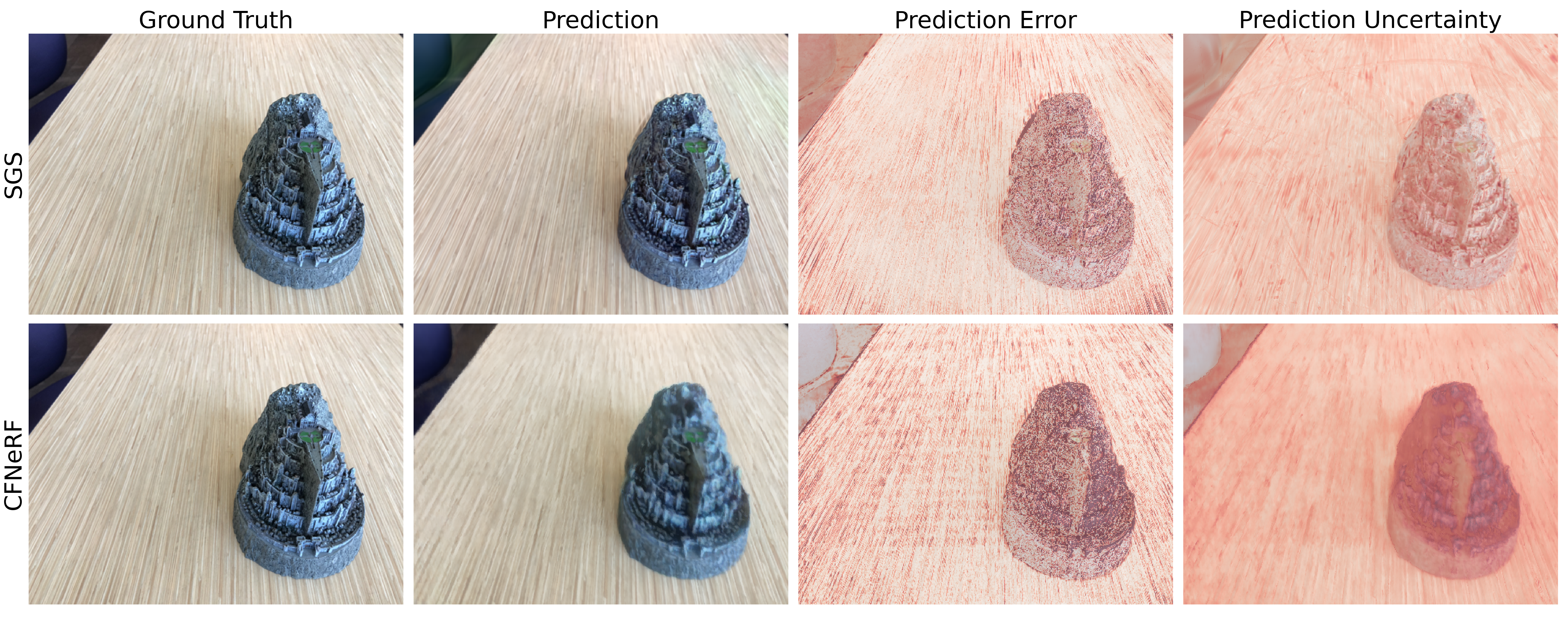}
    \caption{A qualitative example of our method SGS with CF-NeRF \cite{shen2022conditional}. The last column is a visualization of the predicted uncertainty map.}
    \label{fig:qualitative}
\end{figure*}

\subsection{Experimental Setting}

This work addresses the task of synthesizing novel views and associated uncertainty maps from multiple posed views of a static scene. These views consist of RGB photos captured from various camera positions and orientations. The spatial positions of the cameras are referred to as extrinsic parameters, while the intrinsic parameters include the camera's focal length and central point. Similar to prior research, we assume that both extrinsic and intrinsic camera parameters are known for all input views. The objective is to perform standard novel view synthesis using the provided views while also generating an uncertainty map for each synthesized view. Consequently, for every pixel in the novel view, both its color and uncertainty are predicted. It is crucial that the predicted uncertainty correlates with the true error.

We remark that this is the first paper addressing uncertainty estimation for the GS setting, while current literature focuses on NeRF models. This makes direct comparisons challenging, as both absolute image quality as well as metrics for measuring the effectiveness of uncertainty estimation need to be reported. In particular, we remark that the AUSE metric benchmarks uncertainty estimation against the oracle method, so it is relatively insensitive to the absolute image quality generated by each method.

We compare our proposed method with the current literature on NeRF: the state-of-the-art CF-NeRF \cite{shen2022conditional}, the pioneering work of S-NeRF \cite{shen2021stochastic}, and also with NeRF-W \cite{martin2021nerf}, Deep-Ensembles (D.E.) \cite{lakshminarayanan2017simple} and MC-Dropout \cite{gal2016dropout}, as done in previous works. We remark that some methods \cite{shen2021stochastic, shen2022conditional} use extra information in the form of depth maps while we do not, resulting in a setting that is slightly unfair towards SGS. Nevertheless, we report improvements over such methods.

As common practice, all the experiments are conducted on the LLFF dataset from the original NeRF paper \cite{mildenhall2021nerf}, which is composed of eight scenes (\textit{fern}, \textit{flower}, \textit{fortress}, \textit{horns}, \textit{leaves}, \textit{orchids}, \textit{room} and \textit{trex}), using the standard train-test split (e.g., the test split for \textit{fern} is \{\textit{IMG4026}, \textit{IMG4034}, \textit{IMG4042}\}). All the experiments are performed at $\frac{1}{8}$ of the original resolution, so that all synthesized images and uncertainty maps are composed by $504 \times 378$ pixels. 

The hyperparameters in the final loss function are: $\lambda_{\textrm{KL}} = 10^{-3}$, $\lambda_{\textrm{AUSE}} = 5$, $\lambda_{\textrm{ssim}} = 0.2$. For the first $2500$ iterations, we use the standard GS method with the following hyperparameters: the highest spherical harmonics degree is $1$, the densification step is applied until iteration $1000$, and the learning rate of the gaussians centers is fixed to $10^{-2}$.

At iteration $2500$, the current learned GS is fixed and taken as the prior, and the Bayesian regime is introduced. All the prior covariance matrices are initialized as $10^{-2} \mathbb{I}$, where $\mathbb{I}$ is the identity matrix of the correct dimension. The learning rate for all the posterior learnable parameters is set to $10^{-4}$. Then we continue the training until iteration $10000$. During both training and testing, Monte Carlo sampling from the posterior is performed for $8$ times.

\subsection{Main Results}

Table \ref{tab:results} reports our results on the LLFF dataset, evaluating the quality of the rendered images, as well as the reliability of the associated uncertainty maps. The rendered images are quantitatively evaluated with three metrics: PSNR as a distortion metric and LPIPS and SSIM as perceptual metrics. It can be noticed that our method improves by a large margin all these metrics, hence our method has a much lower test prediction bias with respect to previous work.
The uncertainty maps are quantitatively evaluated with two metrics: AUSE RMSE and AUSE MAE. Both metrics are obtained as the area under the sparsification curve, but considering as the error metric the Root Mean Square Error and the Mean Absolute Error, respectively.
As shown in Table \ref{tab:results}, our method improves the AUSE RMSE metric, while keeping an AUSE MAE metric comparable with the state of the art, which however also exploits depth information.
Fig. \ref{fig:qualitative} shows a qualitative result for a novel view generated by SGS and CF-NeRF together with the predicted uncertainty maps. We can notice that SGS is capable of producing a sharp view as well as prediction uncertainty which correlates well with the true rendering error.

\subsection{AUSE Loss Ablation}

\begin{table}[t]
\centering
\caption{AUSE Loss Term ablation.}
\label{tab:auseablation}
\setlength{\tabcolsep}{3pt}
\begin{tabular}{lcccc}
                                                    & PSNR (dB) $\uparrow$   & SSIM $\uparrow$   & LPIPS $\downarrow$& AUSE RMSE $\downarrow$\\ \hline
No AUSE Loss                                        & \textbf{26.65}    & \textbf{0.869}    & \textbf{0.082}    & 0.0291            \\
\textbf{SGS (Ours)}                                          & 24.20             & 0.842             & 0.121             & \textbf{0.0147}   \\ \hline \\
\end{tabular}
\end{table}

Table \ref{tab:auseablation} compares our SGS method with an ablated version, where the proposed AUSE loss term $\mathcal{L}_{\textrm{AUSE}}$ in equation \eqref{eq:lossfinal} is removed in order to verify its effectiveness. As reported in the table, the removal of this loss term improves the photometric reconstruction (measured by the three quality metrics: PSNR, SSIM, and LPIPS), while deteriorating the model's ability to predict accurate uncertainty maps. Hence, this ablation study proves that one of our key contribution, that is to incorporate the AUSE loss term, improves the quality of the predicted uncertainty maps. Moreover, the hyperparameter $\lambda_{\textrm{AUSE}}$ provides a natural way to control the impact of this loss term, so that an application-specific trade-off between reconstruction quality and accurate uncertainty prediction can be found for downstream tasks.

\section{Conclusions}

In this paper, we proposed a novel approach for uncertainty estimation in GS-based novel-view synthesis tasks. Leveraging the efficiency and real-time capabilities of GS, we introduced a stochastic extension to the traditional deterministic GS framework. Our method incorporates uncertainty prediction through a Bayesian framework, specifically using VI to learn the parameters of the GS radiance field. The training is further augmented with direct optimization of the AUSE metric to control the tradeoff between reconstruction quality and accuracy of uncertainty estimation.

Experimental results on the LLFF dataset demonstrated the effectiveness of our approach. We outperformed state-of-the-art methods in terms of rendering quality metrics, while also improving upon uncertainty estimation metrics. Notably, our work advances the state of the art by being the first to introduce uncertainty estimation for GS-based novel-view synthesis tasks.

\bibliographystyle{IEEEtran}

\end{document}